\pdfoutput=1

\documentclass[11pt]{article}

\usepackage[preprint]{acl}

\usepackage{times}
\usepackage{latexsym}
\usepackage{makecell}
\usepackage[T1]{fontenc}

\usepackage[utf8]{inputenc}
\usepackage{amsmath} 
\usepackage{url}

\usepackage{subcaption}
\usepackage{graphicx}
\usepackage{subcaption}
\usepackage{stfloats} 
\usepackage{microtype}
\usepackage{graphicx}
\usepackage{booktabs}
\usepackage{inconsolata}
 \usepackage{tcolorbox}
\usepackage{graphicx}
\usepackage{comment}
\usepackage{tcolorbox}
\usepackage{xcolor}
\definecolor{InkGreen}{rgb}{0.0, 0.39, 0.0}
\title{Diagnosing the Performance Trade-off in Moral Alignment:\\A Case Study on Gender Stereotype}

\author{
    \textbf{Guangliang Liu\textsuperscript{1}\footnotemark[1]}
    ~~\textbf{Bocheng Chen\textsuperscript{2}\footnotemark[1]}
    ~~\textbf{Han Zi\textsuperscript{3}}
    ~~\textbf{Xitong Zhang\textsuperscript{1}}
    ~~\textbf{Kristen Johnson\textsuperscript{1}}
\\
    \textsuperscript{1}Michigan State University
~\textsuperscript{2}University of Mississippi 
~\textsuperscript{3}Northeastern University
\\
\texttt{\{liuguan5,zhangxit,kristenj\}@msu.edu}\\\texttt{bchen5@olemiss.edu}~~~~~\texttt{zi.h@northeastern.edu}
}

\begin{document}
\maketitle
\begin{abstract}
Moral alignment has emerged as a widely adopted approach for regulating the behavior of pretrained language models (PLMs), typically through fine-tuning on curated datasets.
Gender stereotype mitigation is a representational task within the broader application of moral alignment.
However, this process often comes at the cost of degraded downstream task performance.
Prior studies commonly aim to achieve a performance trade-off by encouraging PLMs to selectively forget only stereotypical knowledge through carefully designed fairness objective, while preserving their language modeling capability (overall forgetting).
In this short paper, we investigate whether the performance trade-off can be achieved through the lens of forgetting and the fairness objective.
Our analysis shows that the large datasets needed for satisfactory fairness highlight the limitations of current fairness objectives in achieving an effective trade-off:
(1) downstream task performance is strongly correlated with overall forgetting; 
(2) selective forgetting reduces stereotypes, but overall forgetting increases.
and (3) general solutions for alleviating forgetting are ineffective at reducing the overall forgetting and fail to improve downstream task performance.
~\textit{\footnotesize \textbf{Warning}: examples in this paper are stereotypical and may be offensive.}
\end{abstract}

\section{Introduction}
Moral alignment~\cite{gabriel2020artificial,tennant2024moral} involves fine-tuning or model editing pretrained language models (PLMs) on datasets designed to encourage their morally appropriate behaviors. Reinforcement Learning from Human Feedback (RLHF;~\citet{ouyang2022training}) and Direct Preference Optimization (DPO;~\citet{rafailov2023direct}) are two representative methods within this broader effort.
However, moral alignment often compromises the helpfulness of PLMs, motivating the research community to explore better balance in the helpfulness–harmlessness trade-off~\cite{bai2022constitutionalaiharmlessnessai,milliere2023alignment,chehbouni2025beyond,pelaez2025domain}.

Among moral alignment tasks, mitigating gender stereotypes stands out as a central focus~\cite{kotek2023gender}.
This is not only because gender fairness is a critical societal concern but also because 
the popular moral alignment approach based on reinforcement learning does not satisfactorily achieve gender fairness~\cite{eloundou2024first,wolf2024fundamental,zhou2023lima}.
In this paper we focus on the gender stereotype mitigation task as the causality of gender stereotypes is well defined in the literature, enabling us to have controllable experiments and precise analysis~\cite{liu2025discourse}.
In the context of gender stereotypes mitigation, the trade-off between downstream task performance (helpfulness of PLMs) and fairness (harmlessness of PLMs) is widely recognized~\cite{liang2021towards,liu-etal-2024-towards-understanding,gallegos2024bias}.

When it comes to fine-tuning, one prominent challenge is \textit{catastrophic forgetting}~\cite{kirkpatrick2017overcoming,kemker2018measuring,liu2023pac}: in the course of fine-tuning, models tend to forget knowledge acquired during pre-training.
This is more serious for PLMs, an instance of over-parameterized neural models~\cite{liu2023pac}. 
One representative example of forgetting in the context of gender stereotype mitigation is: PLMs may relearn social stereotypes from downstream tasks, even after prior mitigation efforts~\cite{liu-etal-2024-towards-understanding}.
Previous studies have leveraged the forgetting property of PLMs to encourage the selective forgetting of stereotypical knowledge while preserving the other knowledge, aiming for a better trade-off between fairness and downstream task performance~\cite{webster2020measuring, ilharco2022editing, yu-etal-2023-unlearning, dige2024can}.
In this paper, we define selective forgetting as the bias level in PLMs, measured by the StereoSet score~\cite{nadeem2021stereoset} (closer to 50 indicates better forgetting). Overall forgetting is assessed via language modeling performance, with higher perplexity indicating greater forgetting.

However, whether this trade-off can be effectively achieved remains underexplored, and few studies have systematically examined the effectiveness of existing approaches targeting this trade-off. 
We focus on the trade-off between downstream task performance and fairness through the lens of forgetting and the fairness objective.
As shown in Figure~\ref{fig:motivation}, we adopt counterfactual data augmentation (CDA;~\citet{webster2020measuring}), which flips the gender pronoun to construct an anti-stereotypical corpus, as the method for mitigating gender stereotypes.
Therefore, \textit{the fairness objective primarily focuses on flipping gendered words while overlooking other contextual content in the corpus}.

Our experimental results make four key observations: (1) fairness is influenced by both forgetting and the fairness objective; (2) downstream task performance is highly correlated to the overall forgetting; (3) selective forgetting cannot help reducing the level of overall forgetting; and (4) general approaches to mitigating forgetting are ineffective at reducing overall forgetting and instead degrade downstream task performance.
\section{Related Works\label{app:background}}
Moral alignment has become the most popular approach to regulate the behavior of LLMs; RLHF~\cite{ouyang2022training} is one example method. However,~\citet{eloundou2024first} shows that even when alignment methods are applied to pretrained LLMs, ChatGPT, for example, still exhibits social stereotypes. On the other hand, there have been studies criticizing the effectiveness of current moral alignment methods based on human preference.~\citet{zhi2024beyond} claims that the preference-based representation of morality may not be effective.~\citet{zhou2023lima}, \citet{lin2023unlocking}, and \citet{qi2024safety} empirically demonstrate that existing alignment methods often operate at a superficial level, leaving LLMs vulnerable to attacks. One explanation is that existing alignment methods cannot achieve intrinsic alignment~\cite{wolf2024fundamental}, which is in line with the conclusion of superficial alignment~\cite{zhou2023lima,lin2023unlocking,qi2024safety}. In this paper, we leverage Counterfactual Data Augmentation (CDA)~\cite{webster2020measuring}, which directly modifies the association between gender tokens and attribute tokens, rather than relying on the incidental effects~\cite{roth2017incidental} seen in preference-based alignment methods.

Though the trade-off between fairness and task-specific performance has been a widely recognized challenge~\cite{liang2021towards,gallegos2024bias}, there are few studies which explore the underlying mechanism of this trade-off.~\citet{liu-etal-2024-towards-understanding} investigates the mechanism of the bias transfer hypothesis, but the cause of degraded language modeling remains a mystery. In this paper, we frame both language modeling degradation and social bias mitigation through the lenses of forgetting~\cite{li2024examining}, flipped gender words~\cite{webster2020measuring}, and overexposed context, revealing the mechanism of the trade-off and proposing a solution to alleviate the degraded language modeling by addressing overexposed context. 
\section{Preliminary\label{sec:preliminary}}
\begin{figure}[t]
    \centering
    \includegraphics[width=0.99\linewidth]{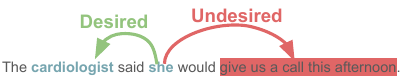}
    \caption{Fairness Objective Motivation. Application of CDA creates the \textcolor{InkGreen}{desired} association between occupation and gender resulting in an anti-stereotypical corpus. However, such fine-tuning introduces the \textcolor{red}{undesired} association of the gender with a neutral phrase. 
    }
    \label{fig:motivation}
\end{figure}

\textbf{Motivation.} According to the dynamic semantics theory in linguistics~\cite{heim2002file,li2021implicit}, a sentence encodes multiple layers of information state, with stereotypes being only one aspect. 
Also, morality operates at the pragmatic level but existing fairness objective simplifies it as a semantic level task~\cite{liu2025diagnosing}.
Figure~\ref{fig:motivation} illustrates the fairness objective adopted in prior studies, which encourages PLMs to associate an occupation, e.g., \textit{cardiologist}, with \textit{she} while forgetting the pretrained association with \textit{he}. 
On one hand, this fairness objective may inadvertently reinforce associations between gendered pronouns and otherwise neutral phrases, such as give us a call this afternoon. And swapping gendered pronouns simplifies a pragmatics-level task.
Due to this undesired association and simplification, we hypothesize that the resulting trade-off in previous studies may not align with expectations.

\textbf{Experimental Setting.} We adopt CDA as a gender stereotype mitigation method for the key reasons: (1) CDA is the prototype for representative mitigation methods based on either model editing or fine-tuning, and (2) conventional alignment\footnote{We refer to conventional alignment as the alignment methods leveraging reinforcement learning.} methods have faced criticism for their inability to achieve intrinsic\footnote{We get gender bias scores (StereoSet) of llama-3.2-1b-base and llama-3.2-1b-instruct. The instruct model achieves a score of 69.196, which is only slightly lower than the base model’s 71.834.} alignment~\cite{wolf2024fundamental}, but CDA directly addresses social bias by manipulating social group identifiers and associated attributes (e.g., occupations).
In terms of backbone models, we utilize BERT-base~\cite{devlin2019bert} and Llama-3.2-1B~\cite{dubey2024llama} to cover different model architectures. 
Notably, we focus on computationally intensive fine-tuning to characterize the performance trade-off. 

We adopt the StereoSet score~\cite{nadeem2021stereoset} as the gender stereotypes score.
We leverage the sentiment analysis task (SST) from the GLUE benchmark~\cite{wang-etal-2018-glue} as the downstream task.
Unless otherwise specified, we fine-tune PLMs using 10K samples from each dataset, as we find that an anti-stereotypical corpus of this size is sufficient to achieve fairness comparable to that obtained with larger datasets.
For more details about the experimental settings and hyperparameters for fine-tuning, please refer to Appendix~\ref{app:setting}.

\textbf{Notations.} Assume we have a news dataset\footnote{https://data.statmt.org/news-commentary/v15/} $\{\mathcal{D}_g, \mathcal{D}_n\}$, where $\mathcal{D}_g$ denotes the subset containing gender-specific words, and $\mathcal{D}_n$ denotes the subset without any gender-specific words.
By following~\citet{zhao-etal-2018-learning} and \citet{liu-etal-2024-towards-understanding}, we flip gendered words in $\mathcal{D}_g$ and obtain a modified dataset $\mathcal{D}_f$ to mitigate gender stereotypes via CDA. 
The datasets are available in the anonymized github\footnote{\url{https://anonymous.4open.science/r/LMinFinetuneData-D0A6/}}.

\section{Analysis\label{sec:mechanism}}
\begin{figure*}[t]
  \centering
  
  \begin{subfigure}[b]{0.28\textwidth}
    \includegraphics[width=\linewidth]{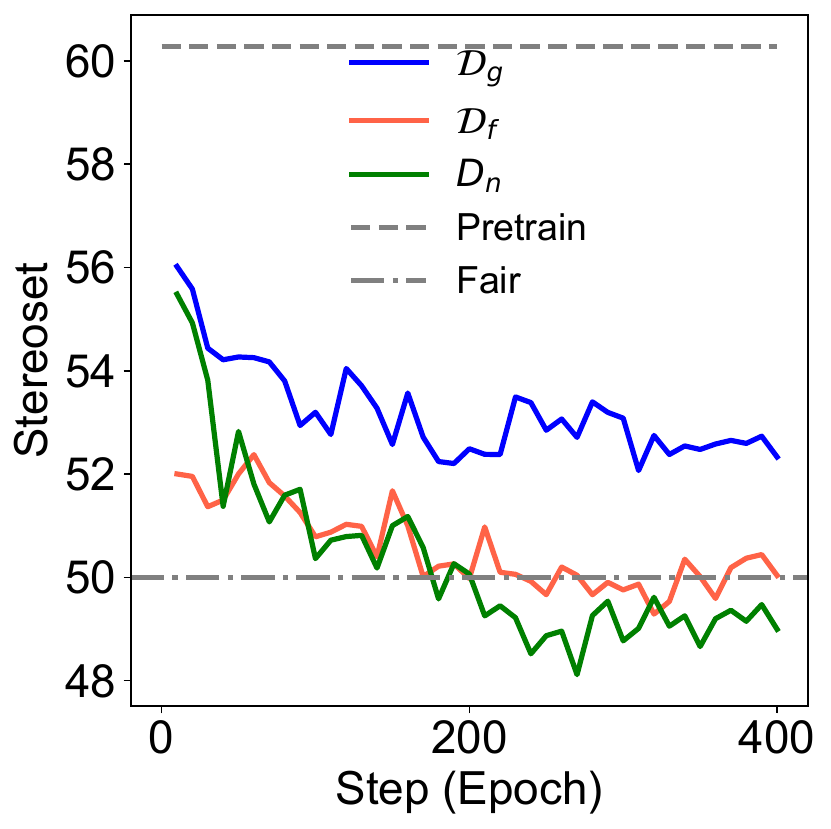}
  \end{subfigure}
  \hfill
  \begin{subfigure}[b]{0.28\textwidth}
    \includegraphics[width=\linewidth]{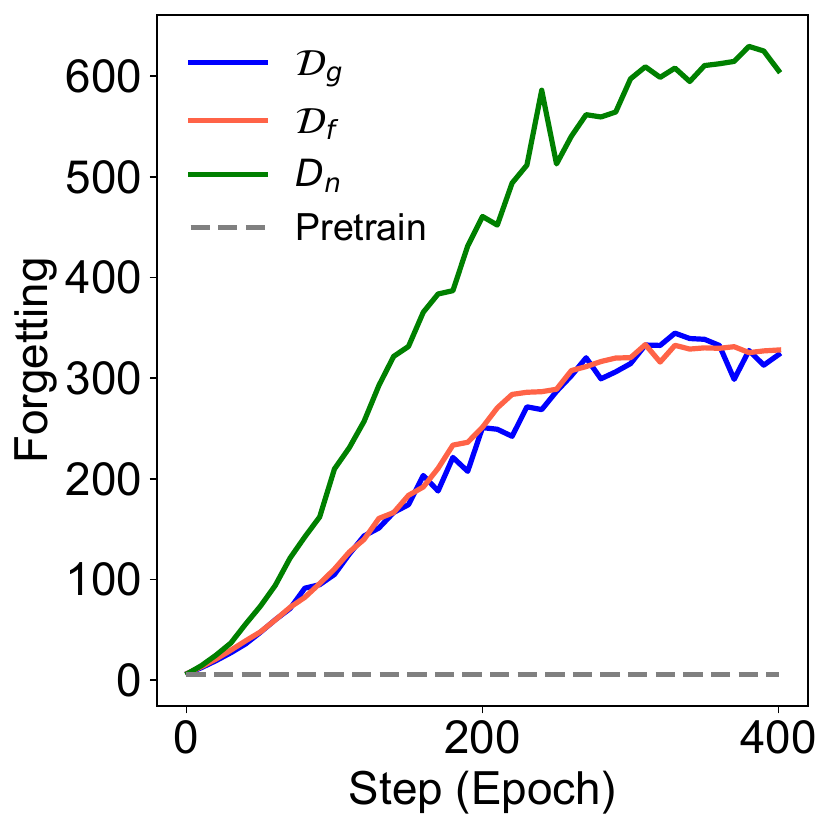}
  \end{subfigure}
    \hfill
  \begin{subfigure}[b]{0.28\textwidth}
    \includegraphics[width=\linewidth]{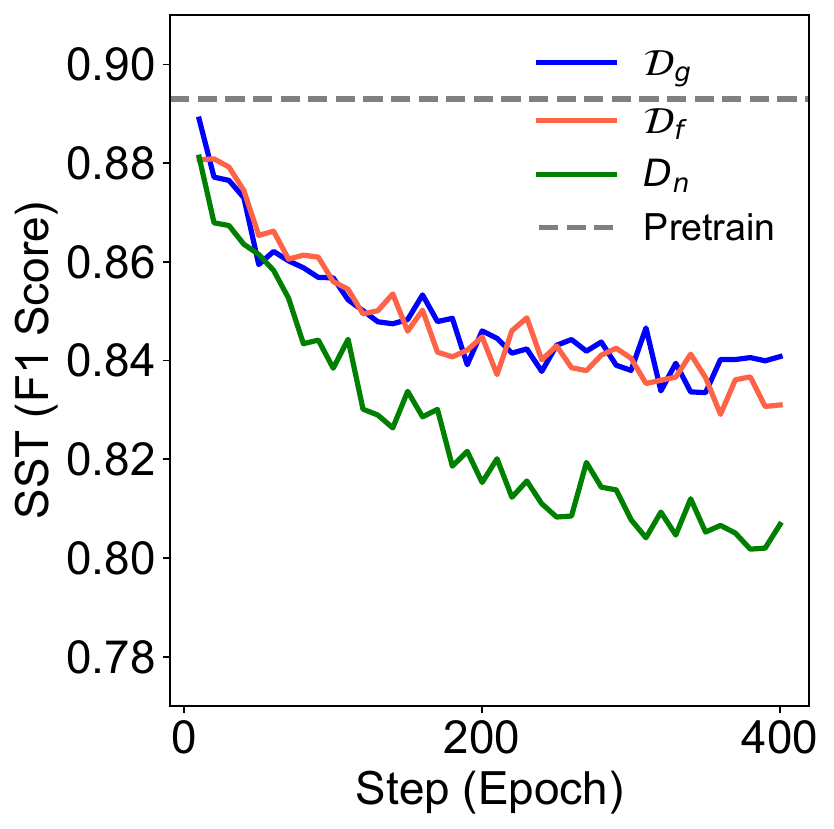}
  \end{subfigure}
  \hfill
  \caption{\small StereoSet Score (\textbf{Left}), Overall Forgetting (\textbf{Middle}) and SST Performance (\textbf{Right}) over Fine-tuning Epochs of BERT. Mechanistic analysis reveals the effects of forgetting and the fairness objective in facilitating gender stereotype mitigation and on downstream SST performance. It is apparent that the StereoSet score is dominated by both forgetting and fairness objective, though the forgetting itself can contribute to satisfactory fairness. The SST performance is governed by forgetting. Additional results for Llama are in Appendix~\ref{app:llama}.}
  \label{fig:mechanism}
\end{figure*}

\begin{figure*}[b]
  \centering
  
  \begin{subfigure}[b]{0.28\textwidth}
    \includegraphics[width=\linewidth]{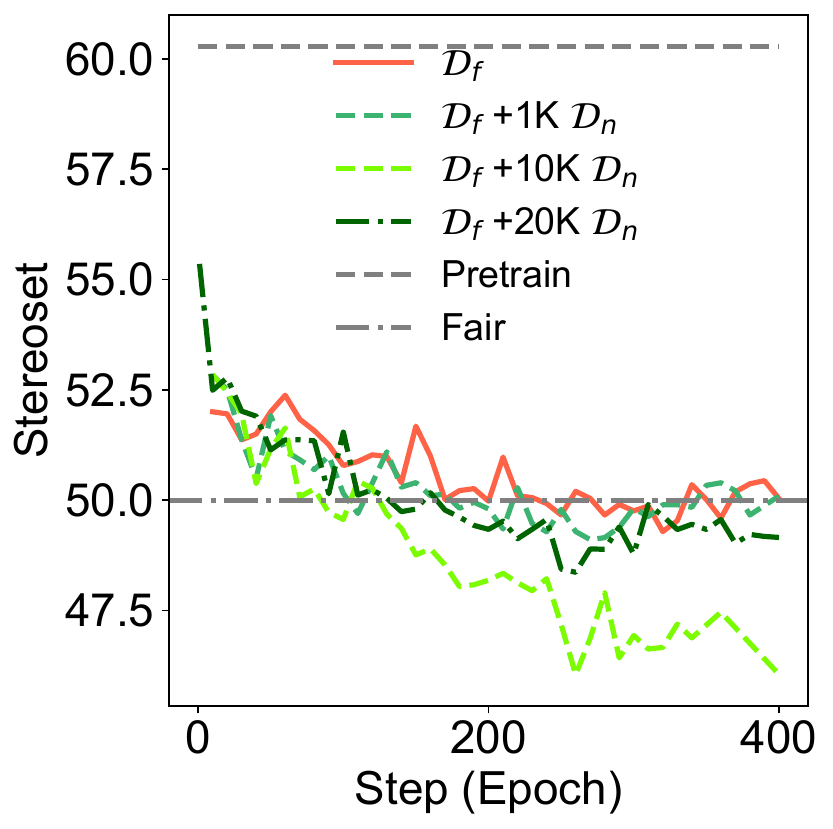}
  \end{subfigure}
  \hfill
  \begin{subfigure}[b]{0.28\textwidth}
    \includegraphics[width=\linewidth]{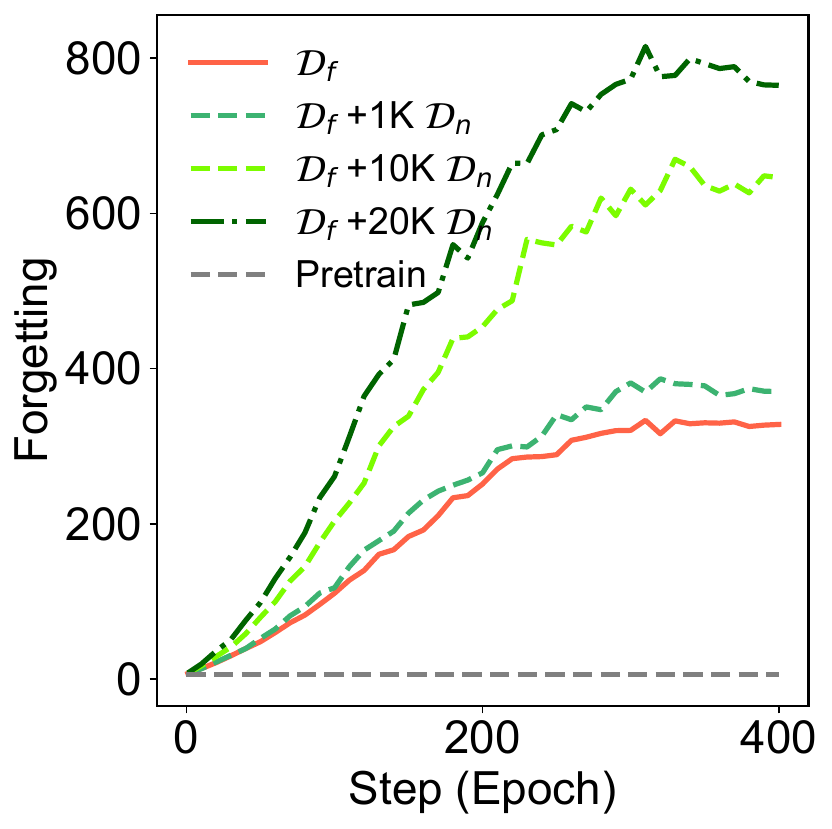}
  \end{subfigure}
    \hfill
  \begin{subfigure}[b]{0.28\textwidth}
    \includegraphics[width=\linewidth]{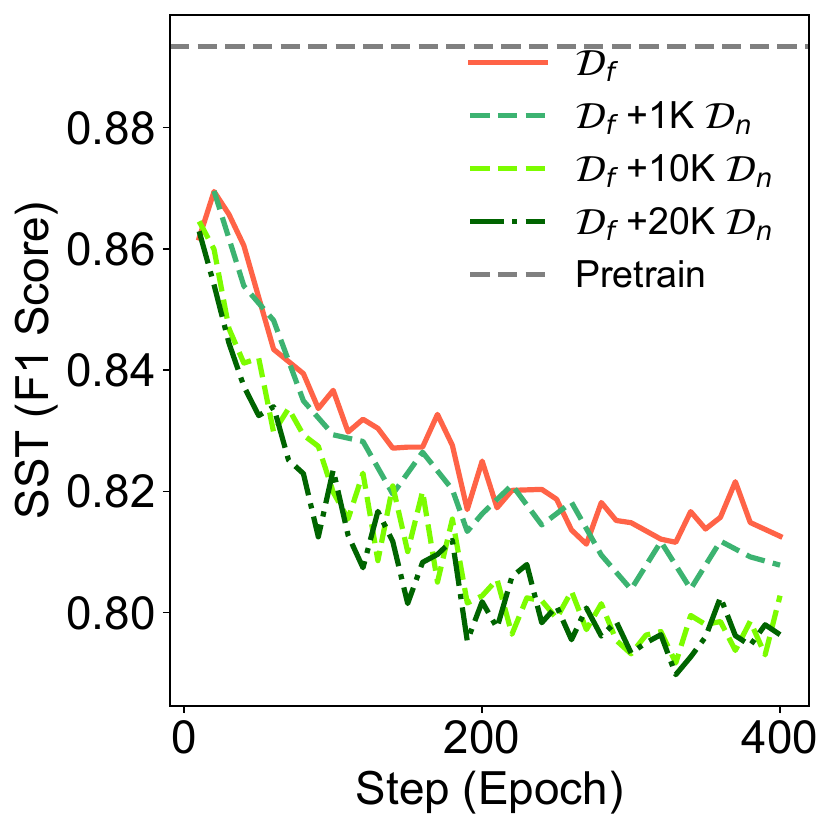}
  \end{subfigure}
  \hfill  
  \caption{\small Experimental results with variant size of fine-tuning dataset with BERT. We take 10K samples from $\mathcal{D}_f$ and consider different size of ${\mathcal{D}_n}$ by following~\citet{webster2020measuring}.~\textbf{Left:} StereoSet score, \textbf{Middle:} Forgetting and \textbf{Right:} SST performance. It is evident that increasing the amount of fine-tuning data leads to greater forgetting and poorer SST performance. Additional results for Llama are in Appendix~\ref{app:llama}.}
    \label{fig:moredata}
\end{figure*}

\begin{figure*}[t]
  \centering
  \hfill
  \begin{subfigure}[b]{0.28\textwidth}
    \includegraphics[width=\linewidth]{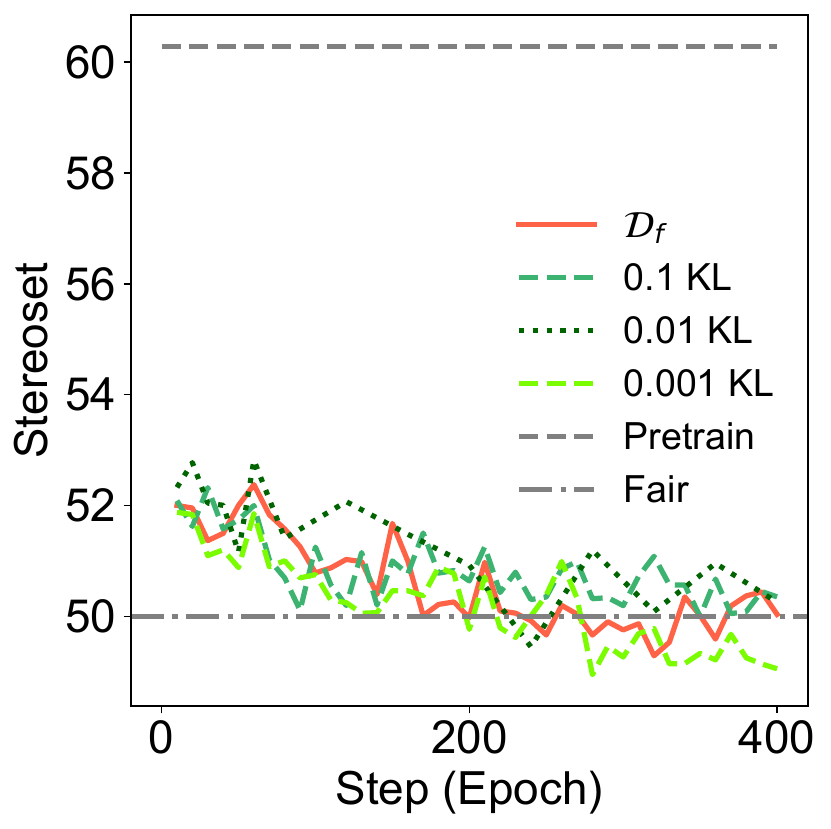}
  \end{subfigure}
  \hfill
  \begin{subfigure}[b]{0.28\textwidth}
    \includegraphics[width=\linewidth]{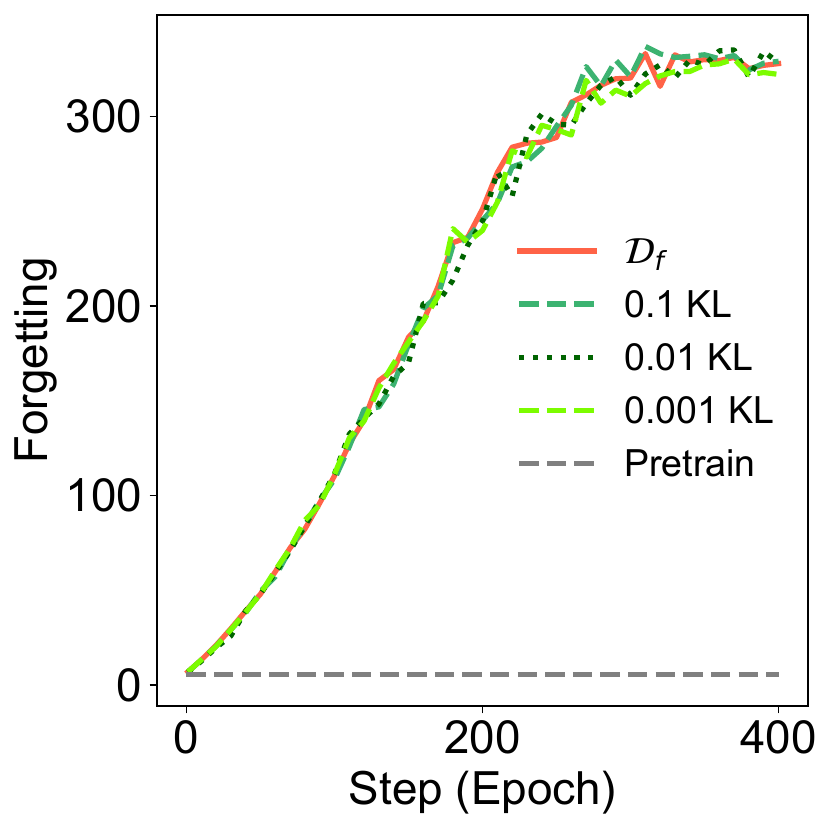}
  \end{subfigure}
    \hfill
  \begin{subfigure}[b]{0.28\textwidth}
    \includegraphics[width=\linewidth]{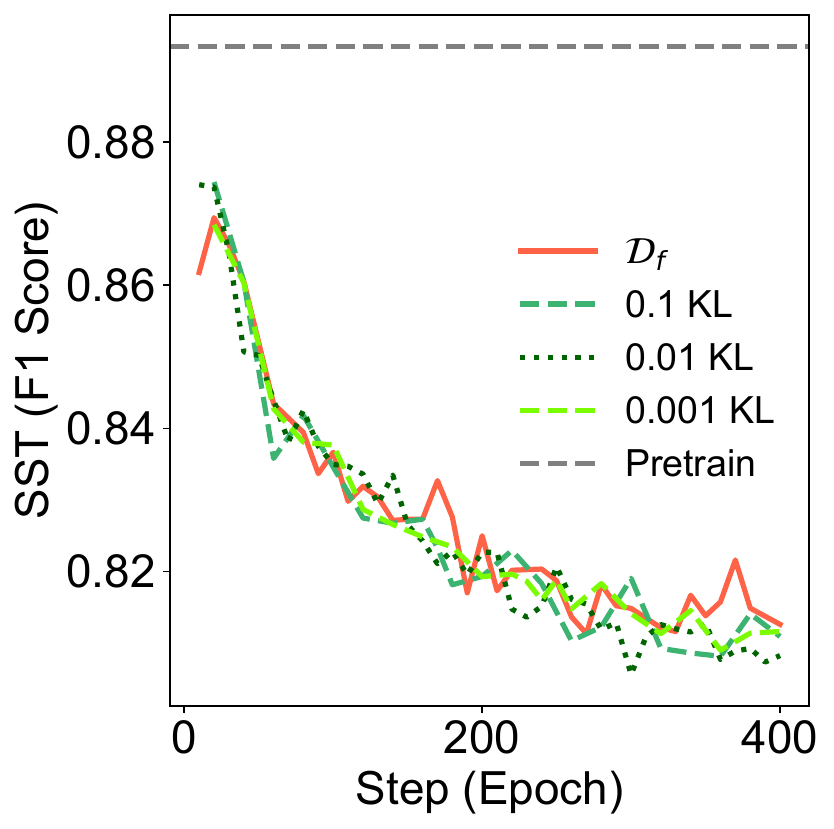}
  \end{subfigure}
  \hfill
   
  \caption{\small Experimental Results with KL-based Regularization for BERT. \textbf{Left:} StereoSet score; \textbf{Middle:} overall forgetting; \textbf{Right:} SST performance. It is apparent the different levels of regularization lead to similar overall forgetting and  SST performance though they caused slight differences in the StereoSet score. Additional results for Llama are in Appendix~\ref{app:llama}.}
    \label{fig:regularization}
\end{figure*}
  
In this section, we analyze how forgetting and the fairness objective influence the trade-off between StereoSet scores (fairness) and SST performance (downstream generalization). Building on this analysis, we further demonstrate the negative effects of general solutions that aim to mitigate forgetting and demonstrate their failure in achieving better trade-off.

\subsection{Selective Versus Overall Forgetting} Figure~\ref{fig:mechanism} shows the experimental results regarding how StereoSet score and SST performance evolve over fine-tuning epochs. 
As shown in the left figure, both $\mathcal{D}_f$ and $\mathcal{D}_n$ enable the model to achieve a StereoSet score of 50, indicating satisfactory fairness. 
Since $\mathcal{D}_n$ contains no gender-specific terms and its only effect is to induce forgetting, we conclude that fairness is correlated to both forgetting and the fairness objective. 
This can be further confirmed by the reduced StereoSet score of fine-tuning PLMs with a stereotypical corpus, $\mathcal{D}_g$.
The converged and stable StereoSet score achieved with $\mathcal{D}_f$ highlights the effectiveness of the fairness objective, whereas $\mathcal{D}_n$ tends to drive PLMs toward inverse stereotypes (e.g., stereotype to females $\rightarrow$ stereotype to males).

Regarding overall forgetting, the middle plot in Figure~\ref{fig:mechanism} shows that both $\mathcal{D}_f$ (anti-stereotypical) and $\mathcal{D}_g$ (stereotypical) contribute equally to the overall forgetting level. 
We attribute this to the unintended associations illustrated in Fig.\ref{fig:motivation}.
For more discussion about the unintended association, please refer to Appendix~\ref{app:genderInNeutral}.
In contrast, $\mathcal{D}_n$ induces significantly greater forgetting, which becomes more pronounced as the amount of $\mathcal{D}_n$ used in fine-tuning increases. 
Interestingly, the SST performance of PLMs closely aligns with their forgetting behavior: models fine-tuned on $\mathcal{D}_f$ and $\mathcal{D}_g$ exhibit comparable SST performance. Whereas greater forgetting leads to substantially degraded SST results, as illustrated by the green lines representing different sizes of $\mathcal{D}_n$.

These observations suggest that downstream task performance is more strongly correlated with overall forgetting, and that the overall forgetting level remains unaffected by the selective forgetting induced by the fairness objective.
A natural question arises: \textit{can we alleviating the overall forgetting while still promoting fairness during fine-tuning, thereby achieving a better trade-off?} Two commonly adopted solutions are (1) augmenting the fine-tuning data with additional corpora~\cite{webster2020measuring,qi2024safety}, and (2) applying a (KL divergence-based) regularization term that penalizes the change of prediction logits to control the learning dynamics~\cite{ouyang2022training,rafailov2023direct}.
We will present evidence that these solutions are ineffective in reducing the overall forgetting.
\subsection{Strategy for Alleviating Forgetting}
\textbf{Additional Corpora.}
Given our findings on the significant role of overall forgetting for downstream SST performance, we hypothesize that this solution should be negative as incorporating more fine-tuning data tends to increase forgetting.
Figure~\ref{fig:moredata} presents the experimental results of fine-tuning the BERT model with $\mathcal{D}_f$ and varying volumes of $\mathcal{D}_n$.
The left subfigure shows that all fine-tuning configurations achieve satisfactory fairness; however, as shown in the middle subfigure, the overall forgetting level consistently increases with the inclusion of more $\mathcal{D}_n$.
The SST performance in the right subfigure follows the same trend as overall forgetting: $\mathcal{D}_f$ yields the best performance, whereas fine-tuning with the largest volume of data results in the worst SST performance.
These observations suggest that simply adding more data during fine-tuning is ineffective, as it amplifies overall forgetting and degrades downstream performance.

\textbf{Regularization.}
A popular method to avoid forgetting too much is to apply a KL divergence regularization term over the PLMs' prediction logits~\cite{ouyang2022training,rafailov2023direct}.
Using this method, we explore multiple regularization coefficients: 0.1, 0.01 and 0.001.
The left subfigure in Figure~\ref{fig:regularization} shows that with different regularization coefficients, PLMs can achieve satisfactory fairness.
However, fine-tuning with regularization shows comparable overall forgetting as fine-tuning with $\mathcal{D}_f$ alone, which explains why the SST performance with regularization is very close to that of fine-tuning without regularization.
These observations further demonstrate that selectively forgetting stereotypes has little impact on the overall forgetting level, and that applying regularization neither effectively reduces overall forgetting nor improves downstream task performance.

\subsection{Undesired Associations\label{app:genderInNeutral}}
\begin{figure}[ht]
    \centering
    \includegraphics[width=0.8\linewidth]{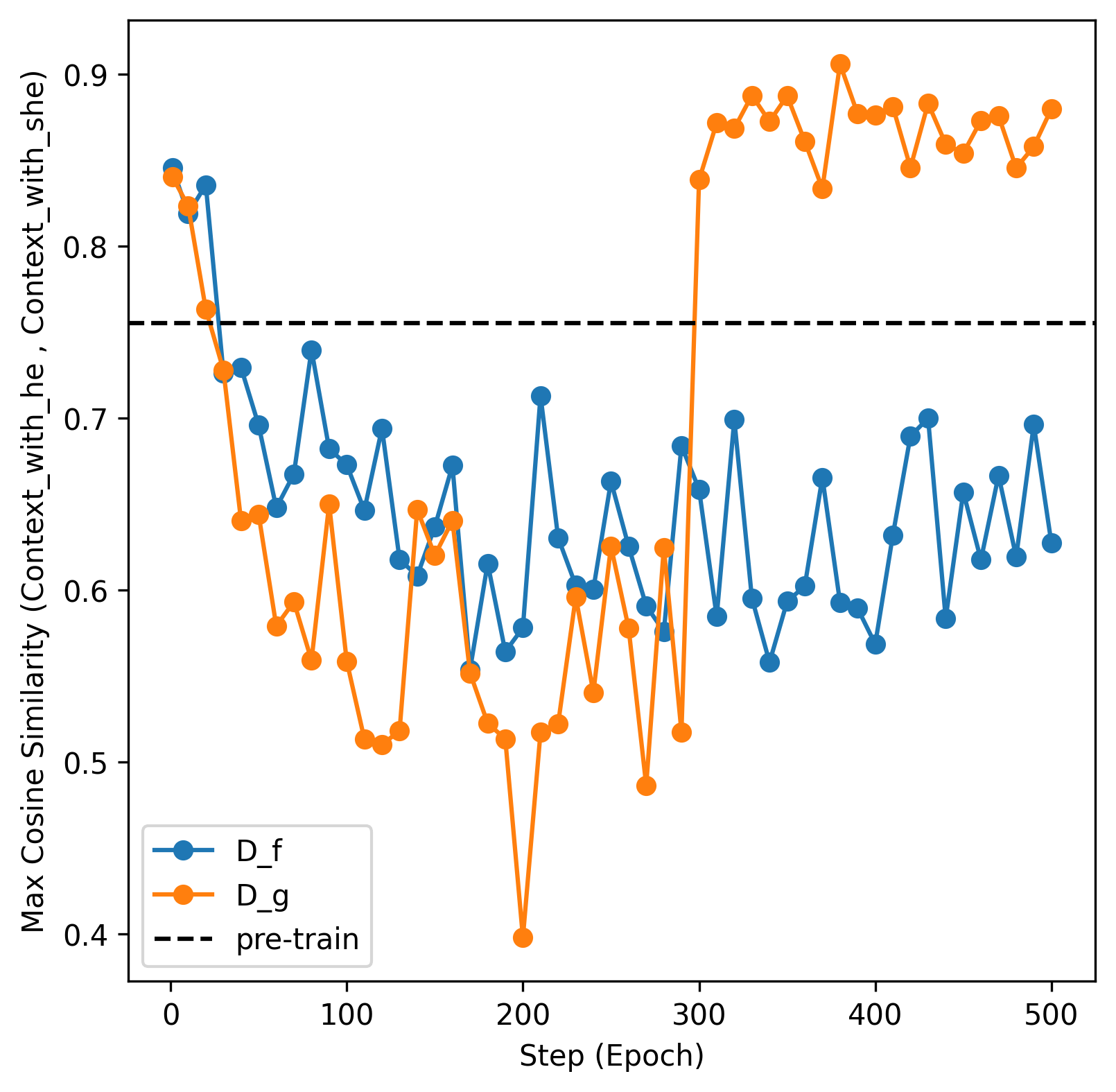}
    \caption{\small Gender Information in Neutral Phrases Acquired from the Debiased BERT Model.}
    \label{fig:genderinfo}
\end{figure}

In this section, we demonstrate that the representations of neutral phrases are affected by undesired associations introduced by the fairness objective. We have shown that the forgetting observed during gender stereotype mitigation, driven by the use of large debiasing datasets, is strongly correlated with degraded downstream generalization. 
Furthermore, we demonstrate, according to our motivation in Figure~\ref{fig:motivation}, that the fairness objective itself introduces noise into the representations of neutral contexts. These findings allow for a more comprehensive methodological investigation of the trade-off problem.

We extract neutral phrases from the Winogender benchmark, e.g., \textit{The technician told the customer that he could [pay with cash]}. The content within [ ] is treated as a neutral context, denoted as $x$. We then compute
$$y = \max(\text{sim}(x, \text{she}), \ \text{sim}(x, \text{he}))$$ where \text{sim} indicates the cosine similarity.
The rationale is that if a PLM is biased, the representation of $x$ will be much closer to either she or he, resulting in a large y. Conversely, if a PLM is successfully debiased, $x$ should be associated with both she and he, producing a smaller $y$, which is the goal of CDA.
In Figure~\ref{fig:genderinfo}, we observe that fine-tuning on $D_f$ consistently reduces $y$ relative to the pretrained checkpoint. However, while $D_g$ initially shows a similar trend, $y$ suddenly increases beyond the pretrained checkpoint. This occurs because the PLM overfitts to the highly stereotypical corpus $D_g$. Overall, it indicates how the fairness objective perturb the representation of neutral contexts.

\textbf{In summary}, 
The downstream task-specific performance is strongly correlated with overall forgetting and general approaches designed to alleviate forgetting often fail to effectively reduce overall forgetting, thereby preventing PLMs from achieving a favorable performance trade-off.
This is expected, as the large debiasing dataset needed for satisfactory fairness inevitably induces forgetting, limiting the success of augmentation and regularization.

\section{Discussion and Future Works\label{sec:discussion}}
The surprisingly significant impact of overall forgetting, which drives the trade-off between helpfulness and harmlessness in PLMs, highlights the limitations of fairness alignment objectives.
The serious forgetting is derived from the large dataset requires for satisfactory fairness, the fundamental reason is that the fairness objective overlook the pragmatic nature of social bias mitigation but employ a semantics-driven solution to solve it.
Based on our findings, we discuss some future works.

\textbf{Pragmatics of Morality.} PLMs capture distributional semantics~\cite{boleda2020distributional} which is reflected by the pretraining objective of token associations, however, gender stereotypes and morals lie in pragmatics~\cite{liu2025diagnosing}.
Therefore, a natural and reasonable hypothesis is that both the used CDA and RL-based alignment methods address a pragmatics-level task with a solution driven by distributional semantics.
For instance, the undesired association arises because the fairness objective focuses solely on social groups and attributes while neglecting the contextual information in which they are embedded.
The most straightforward reason is that this approach requires a large volume of debiasing data to achieve satisfactory fairness performance, therefore it is not surprising that it would cause serious overall forgetting.
We believe a potential solution is to explore how to model the moral alignment problem through a perspective of pragmatics.

\textbf{Fusing debiased models and undebiased models.} Since the trade-off is driven by the inevitable forgetting in the course of gender stereotype mitigation. Therefore, the solution proposed by~\citet{liang2021towards}, which fuses responses from both a debiased and an undebiased model, appears promising. Accordingly, future research could focus on developing effective methods for performing this fusion.

\textbf{Diagnostics.} Diagnosing the effectiveness of moral alignment methods is challenging because: moral alignment serves as an intermediate step between pretraining and downstream task-specific applications, which may limit the extent to which its effects are reflected in the performance of those downstream tasks;
Though there are challenges in diagnosing moral alignment, our findings in this paper is rather different from the general beliefs in the significance of the stereotype mitigation objective~\cite{webster2020measuring,ravfogel2020null}. 
Studies aimed at diagnosing moral alignment methods would make valuable contributions to this research area, particularly given existing evidence highlighting the superficial nature of current alignment approaches.

\textbf{Alignment Tasks.} Moral alignment encompasses a range of tasks, such as non-toxicity, fairness, and robustness to adversarial attacks. Developing a unified diagnostic tool is non-trivial, as these tasks often exhibit distinct linguistic characteristics. For example, in jailbreaking attack defense, the trade-off between safety and over-refusal largely stems from the association between refusal responses and benign content embedded in harmful queries. Moreover, due to the limited size (around 300 samples) of fine-tuning datasets~\cite{qi2024safety}, the effects of forgetting can be particularly subtle and nuanced. Another example is toxicity, which reflects biases in the pretraining corpora. However, unlike stereotypes, defined as disparities between social groups under the same context, toxicity is more implicit and not as easily detectable~\cite{chen2025pragmatic}.

\section{Conclusion\label{sec:conclusion}}
We investigate the trade-off in moral alignment by studying the interaction of forgetting and fairness objectives in gender stereotype mitigation. The downstream performance degradation is more correlated with the overall forgetting, selective forgetting is effective for mitigating social bias but has little to no correlation to downstream performance, and general mitigation strategies can not alleviate overall forgetting, indicating that current fairness objectives limit better trade-offs.

\section{Limitations}
In this paper, we explore the helpfulness-harmlessness trade-off in moral alignment, with an emphasis on the gender stereotypes mitigation task. Though this task is very challenging, further analysis can be extended to other moral alignment tasks such as toxicity detection and jailbreaking defense.
Furthermore, due to the high computational cost, we restrict our analysis to models with up to 1 billion parameters. Future empirical studies could incorporate larger models.
\bibliography{custom}
\appendix

\section{Appendix}
\label{sec:appendix}

\subsection{Experimental setting\label{app:setting}}
See Table~\ref{tab:optimhyperparam}.
\begin{table}[h]
    \centering
    \begin{tabular}{cc}
    \toprule
     \textbf{Hyperparameters} & \textbf{Setting} \\
     \midrule
     \textbf{Optimizer} & \text{AdamW}\\
     \textbf{Learning rate for $\theta$} & \text{5e-5} \\
     \textbf{Maximum training epochs} & \text{400} \\
     \textbf{Batch size} & \text{16}\\
     \bottomrule
    \end{tabular}
    \caption{Hyperparameter Settings for the AdamW Optimizer.}
    \label{tab:optimhyperparam}
\end{table}

\subsection{Additional Results for Llama\label{app:llama}}
\begin{figure*}[h]
  \centering
  \begin{subfigure}[b]{0.28\textwidth}
    \includegraphics[width=\linewidth]{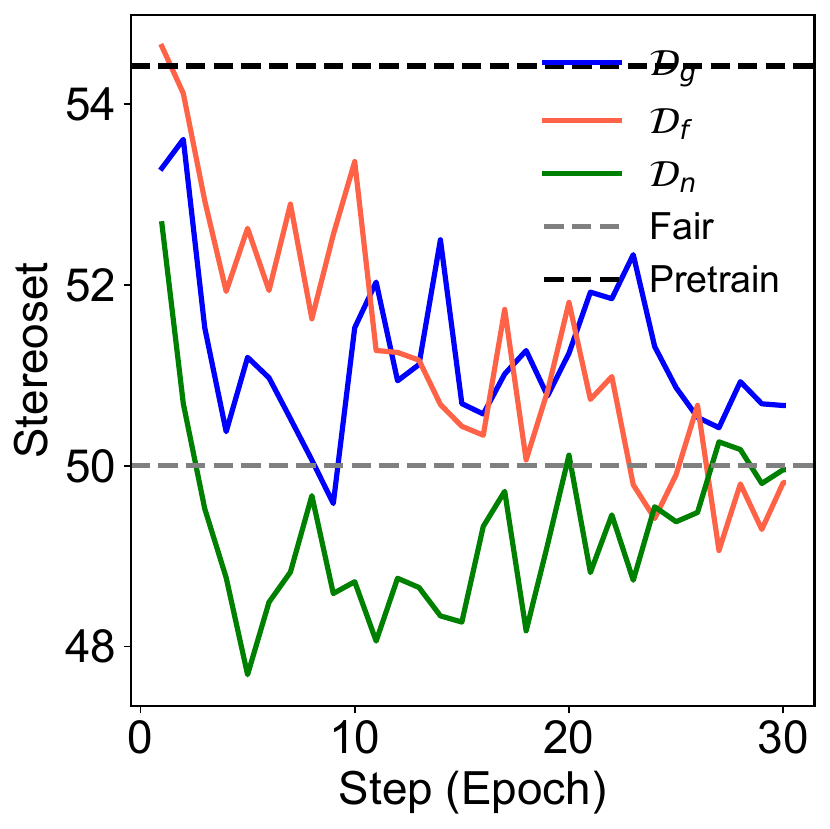}
  \end{subfigure}
  \hfill
  \begin{subfigure}[b]{0.28\textwidth}
   \includegraphics[width=\linewidth]{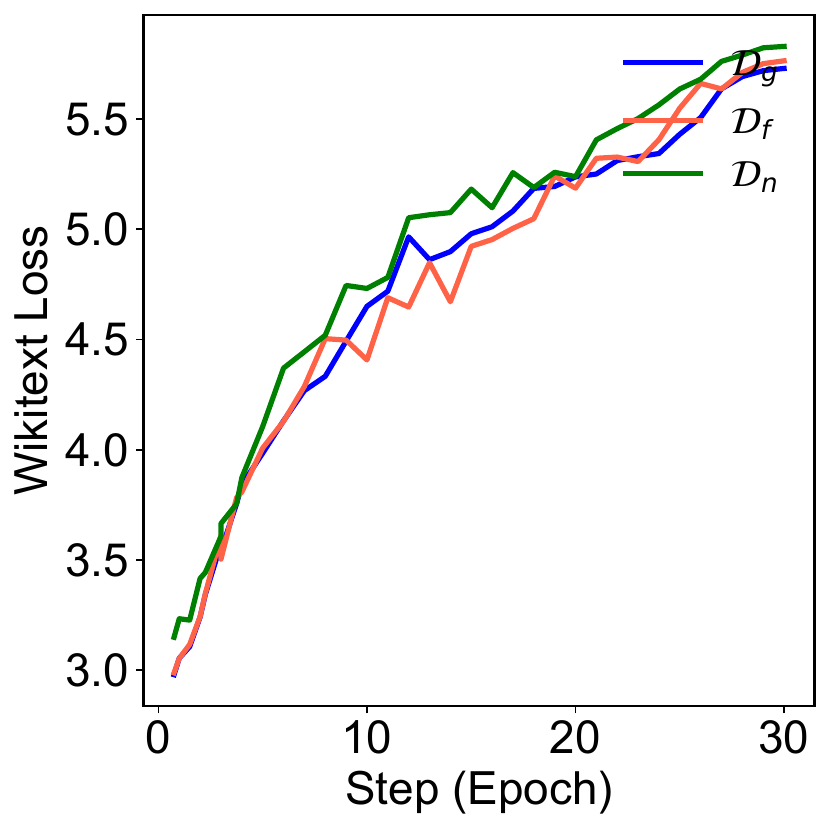}
  \end{subfigure}
   \hfill
  \begin{subfigure}[b]{0.28\textwidth}
   \includegraphics[width=\linewidth]{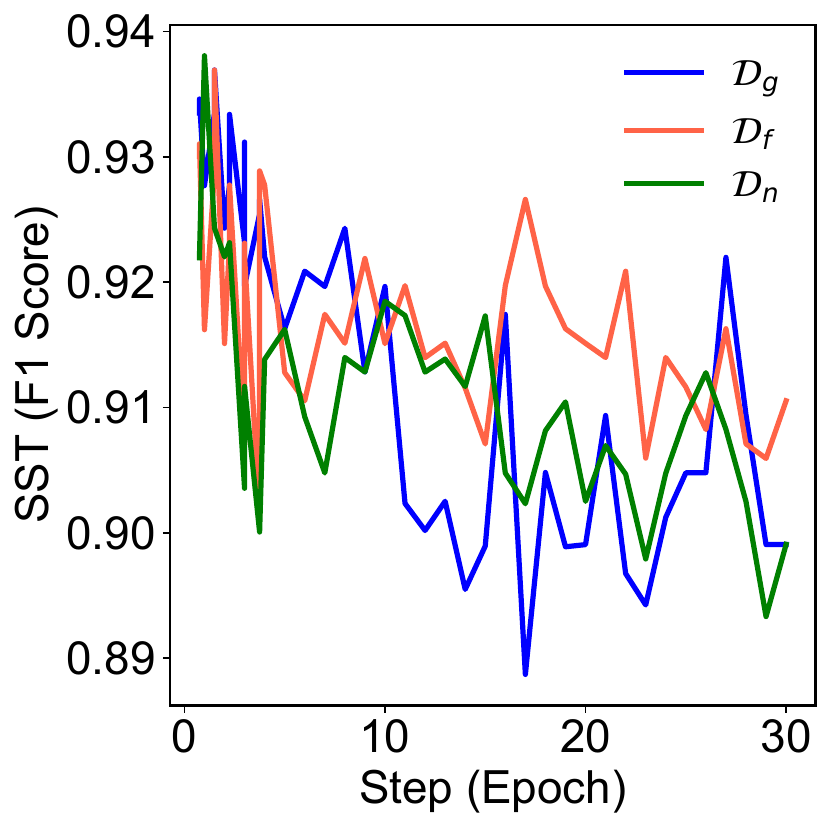}
  \end{subfigure}
  \hfill  
  \caption{\small StereoSet Score (\textbf{Left}), Forgetting (\textbf{Middle}) and SST Performance (\textbf{Right}) over Fine-tuning Epochs of Llama. }
\label{fig:forgetting4llama}
\end{figure*}

\begin{figure*}[h]
  \centering
  \begin{subfigure}[b]{0.28\textwidth}
    \includegraphics[width=\linewidth]{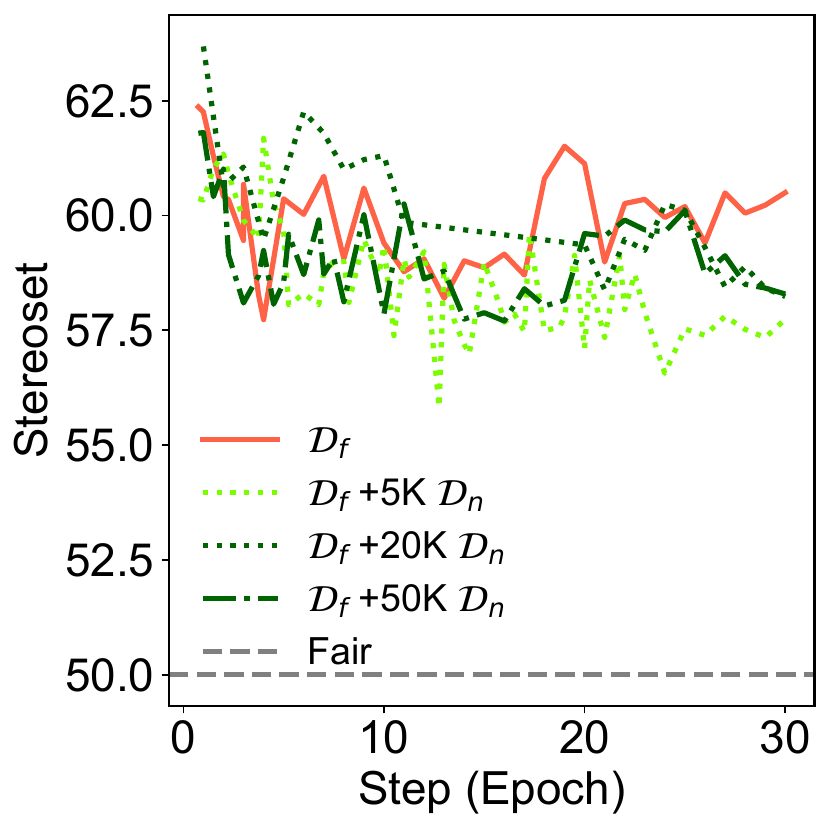}
  \end{subfigure}
  \hfill
  \begin{subfigure}[b]{0.28\textwidth}
   \includegraphics[width=\linewidth]{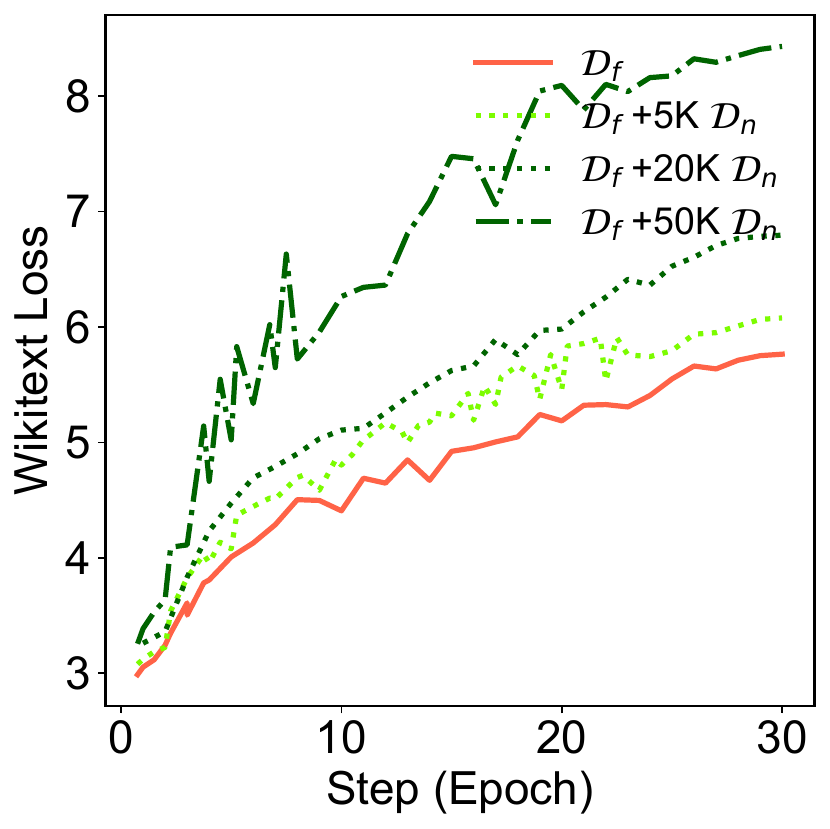}
  \end{subfigure}
   \hfill
  \begin{subfigure}[b]{0.28\textwidth}
   \includegraphics[width=\linewidth]{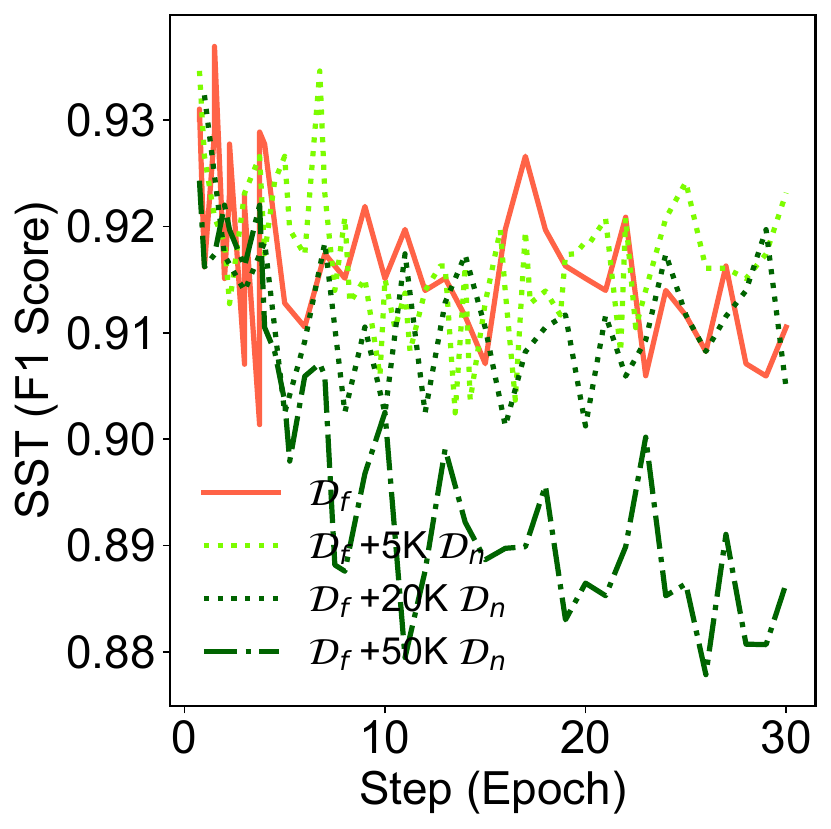}
  \end{subfigure}
  \hfill  
  \caption{\small Experimental results with variant size of fine-tuning dataset with Llama. We take 10K samples from $\mathcal{D}_f$ and consider different size of ${\mathcal{D}_n}$ by following~\citet{webster2020measuring}.~\textbf{Left:} StereoSet score, \textbf{Middle:} Forgetting and \textbf{Right:} SST performance. }
\label{fig:dataaug4llama}
\end{figure*}

\begin{figure}[h]
  \centering
  \begin{subfigure}[b]{0.234\textwidth}
    \includegraphics[width=\linewidth]{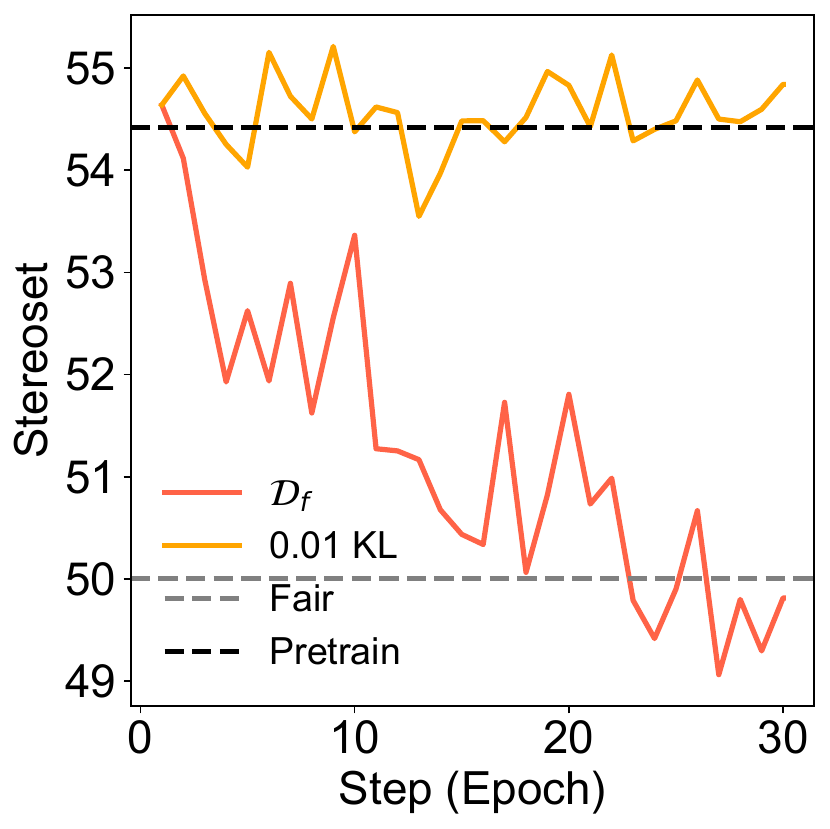}
  \end{subfigure}
  \hfill
  \begin{subfigure}[b]{0.234\textwidth}
   \includegraphics[width=\linewidth]{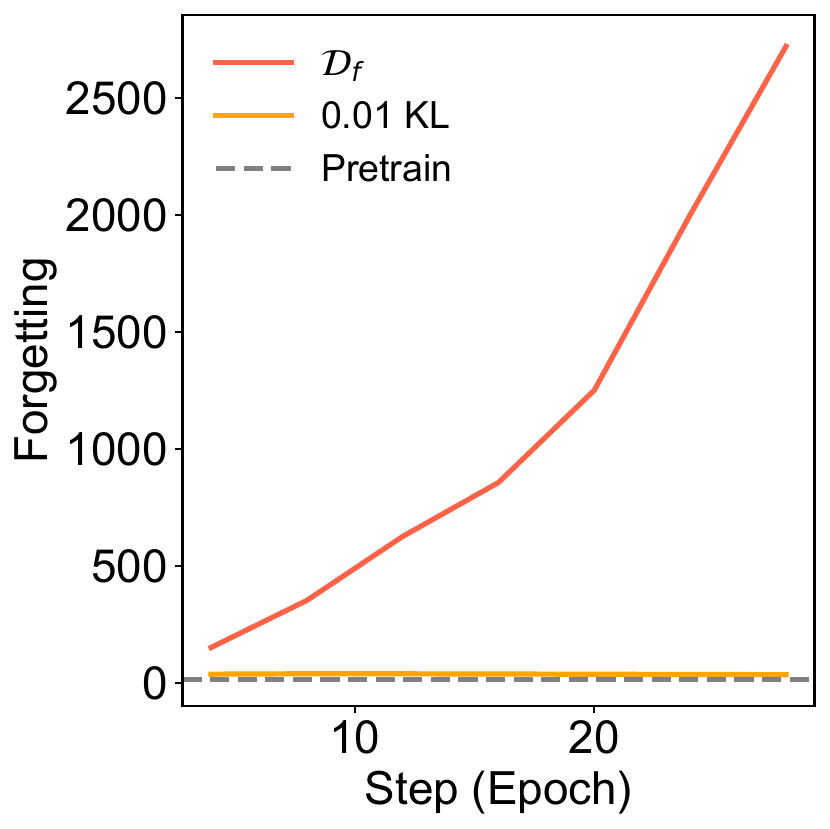}
  \end{subfigure}
  \hfill  
  \caption{\small  Experimental Results with KL-based Regularization for Llama. 
  \textbf{Left:} StereoSet score; \textbf{Right:} overall forgetting; 
  }
\label{fig:regularization4llama}
\end{figure}

Figure~\ref{fig:forgetting4llama} shows the debiasing performance, overall forgetting and SST performance with different fine-tuning datasets $D_f$, $D_n$ and $D_g$.
Unlike our empirical observations for the BERT model, we do not observe significant differences among the three fine-tuning settings, although slight variations exist.
Even the general corpus $D_n$ induces slightly higher forgetting, resulting in SST performance comparable to $D_g$ and $D_f$. 
Nevertheless, as with the BERT model, anti-stereotypical and stereotypical data produce similar overall forgetting.
Figure~\ref{fig:dataaug4llama} presents the effects of data augmentation for the Llama model. 
Similar to what we observe from the BERT model, incorporating more augmentation data can inevitably lead to worse forgetting and worse SST performance.
Figure~\ref{fig:regularization4llama} shows the effects of regularization on the Llama model. 
The regularization prevents LLMs from effectively learning the fairness objective, thereby hindering their ability to achieve fairness, even though it results in a lower perplexity (the pretrained model’s perplexity is 19, while that of the regularized model is 40).
Under these circumstances, evaluating SST performance is not meaningful.

\end{document}